\documentclass[10pt,twocolumn,letterpaper]{article}

\pdfoutput=1

\usepackage{cvpr}
\usepackage{times}
\usepackage{epsfig}
\usepackage{graphicx}
\usepackage{amsmath}
\usepackage{amssymb}
\usepackage{flushend}
\usepackage{color}

\usepackage{algorithm2e}

\usepackage[pagebackref=true,breaklinks=true,letterpaper=true,colorlinks,bookmarks=false]{hyperref}

\usepackage{textcomp}
\usepackage[T1]{fontenc}
\usepackage[latin9]{inputenc}

\newcommand{\comment}[1]{}

\author{ Cl\'ement Farabet$^{1,2}$ \and Camille Couprie$^1$ \and
  Laurent Najman$^2$ \and Yann LeCun$^1$ \and \\ 
  $^1$ Courant Institute of Mathematical Sciences\\ 
  New York University, New York, NY 10003, USA \and \\ 
  $^2$ Universit\'e Paris-Est, Laboratoire d'Informatique Gaspard-Monge\\ 
  \'Equipe A3SI - ESIEE Paris, France
}

\cvprfinalcopy

\def\cvprPaperID{****} 

\makeatother

\begin{document}

\title{Scene Parsing with Multiscale Feature Learning, \\ Purity Trees, and Optimal Covers }

\maketitle

\begin{abstract}
Scene parsing, or semantic segmentation, consists in labeling each
pixel in an image with the category of the object it belongs to. 
It is a challenging task that involves the simultaneous detection,
segmentation and recognition of all the objects in the image.

The scene parsing method proposed here starts by computing a tree of
segments from a graph of pixel dissimilarities. Simultaneously, a set
of dense feature vectors is computed which encodes regions of multiple
sizes centered on each pixel. The feature extractor is a multiscale
convolutional network trained from raw pixels. The feature vectors
associated with the segments covered by each node in the tree are
aggregated and fed to a classifier which produces an estimate of the
distribution of object categories contained in the segment.  A subset
of tree nodes that cover the image are then selected so as to maximize
the average ``purity'' of the class distributions, hence maximizing
the overall likelihood that each segment will contain a single
object. The convolutional network feature extractor is trained
end-to-end from raw pixels, alleviating the need for engineered
features.  After training, the system is parameter free.

The system yields record accuracies on the Stanford Background
Dataset (8 classes), the Sift Flow Dataset (33 classes) and the
Barcelona Dataset (170 classes) while being an order of magnitude
faster than competing approaches, producing a $320\times240$ image
labeling in less than~1 second.
\end{abstract}

\section{Overview}

Full scene labeling (FSL) is the task of labeling each pixel in a
scene with the category of the object to which it belongs. FSL
requires to solve the detection, segmentation, recognition and
contextual integration problems simultaneously, so as to produce a
globally consistent labeling. One of the obstacles to FSL is that the
information necessary for the labeling of a given pixel may come from
very distant pixels as well as their labels. The category of a pixel
may depend on relatively short-range information (e.g. the presence of
a human face generally indicates the presence of a human body nearby),
as well as on very long-range dependencies (is this grey pixel part of
a road, a building, or a cloud?).

\begin{figure}
\begin{centering}
\includegraphics[width=0.99\columnwidth]{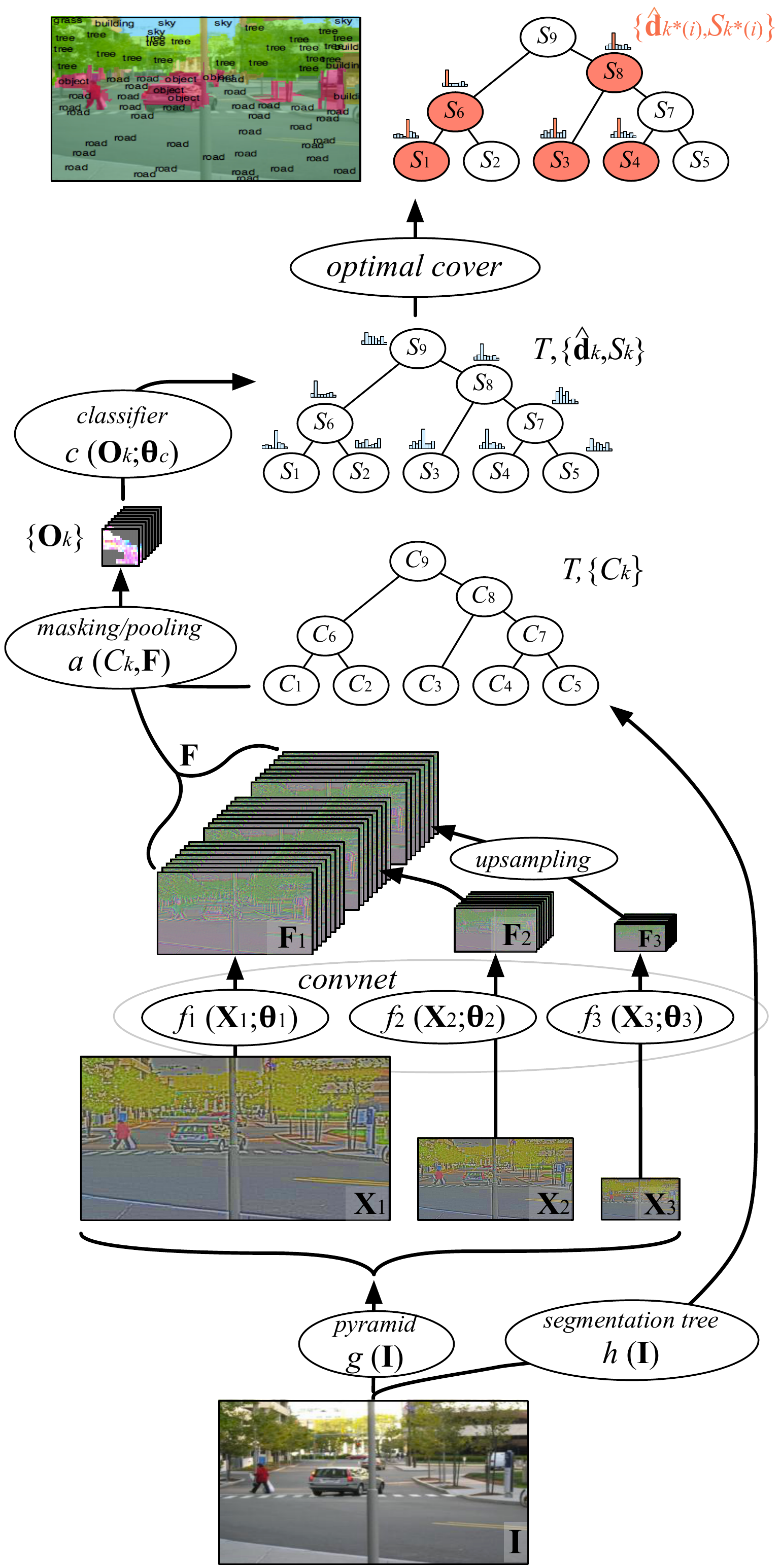}
\end{centering}

\caption{Diagram of the scene parsing system. The raw input image is
  transformed through a Laplacian pyramid. Each scale is fed to a
  2-stage convolutional network, which produces a set of feature
  maps. The feature maps of all scales are concatenated, the
  coarser-scale maps being upsampled to match the size of the
  finest-scale map. Each feature vector thus represents a large
  contextual window around each pixel. In parallel, a segmentation
  tree is computed via the minimum spanning tree of the dissimilarity
  graph of neighboring pixels. The segment associated with each node
  in the tree is encoded by a spatial grid of feature vectors pooled
  in the segment's region. A classifier is then applied
  to all the aggregated feature grids to produce a histogram of
  categories, the entropy of which measures the ``impurity'' of the
  segment. Each pixel is then labeled by the minimally-impure node
  above it, which is the segment that best ``explains'' the pixel.
  \label{fig:model}}
\end{figure}

This paper proposes a new method for FSL, depicted on
Figure~\ref{fig:model} that relies on five main ingredients: {\bf 1)
  Trainable, dense, multi-scale feature extraction}: a multi-scale,
dense feature extractor produces a series of feature vectors for
regions of multiple sizes centered around every pixel in the image,
covering a large context. The feature extractor is a two-stage
convolutional network applied to a multi-scale contrast-normalized
laplacian pyramid computed from the image. The convolutional network
is fed with raw pixels and trained end to end, thereby alleviating the
need for hand-engineered features; {\bf 2) Segmentation Tree}: A graph
over pixels is computed in which each pixel is connected to its 4
nearest neighbors through an edge whose weight is a measure of
dissimilarity between the colors of the two pixels.
A segmentation tree is then constructed using a classical region
merging method, based on the minimum spanning tree of the graph.
Each node in the tree corresponds to
a potential image segment. The final image segmentation will be a
judiciously chosen subset of nodes of the tree whose corresponding
regions cover the entire image.  {\bf 3) Region-wise feature
  aggregation}: for each node in the tree, the corresponding image
segment is encoded by a $5\times5$ spatial grid of aggregated feature
vectors. The aggregated feature vector of each grid cell is computed
by a component-wise max pooling of the feature vectors centered on all
the pixels that fall into the grid cell; This produces a
scale-invariant representation of the segment and its surrounding;
{\bf 4) Class histogram estimation}: a classifier is then applied to
the aggregated feature grid of each node. The classifier is trained to
estimate the histogram of all object categories present in its input
segments; 5) {\bf Optimal purity cover}: a subset of tree nodes is
selected whose corresponding segments cover the entire image. The
nodes are selected so as to minimize the average ``impurity'' of the
class distribution.  The class ``impurity'' is defined as the entropy
of the class distribution. The choice of the cover thus attempts to
find a consistent overall segmentation in which each segment contains
pixels belonging to only one of the learned categories.

All the steps in the process have a complexity linear (or almost
linear) in the number of pixels. The bulk of the computation resides
in the convolutional network feature extractor. The resulting system
is very fast, producing a full parse of a $320\times240$ image in less
than 1 second on a conventional CPU.  Once trained, the system is
parameter free, and requires no adjustment of thresholds or other
knobs.

There are three key contributions in this paper 1) using a {\em
  multi-scale convolutional net} to learn good features for region
classification; 2) using a {\em class purity criterion} to decide if a
segment contains a single objet, as opposed to several objects, or part
of an object; 3) an efficient procedure to obtain a {\em cover that
  optimizes} the overall class purity of a segmentation.

\section{Related work}

The problem of scene parsing has been approached with a wide variety
of methods in recent years. Many methods rely on MRFs, CRFs, or other
types of graphical models to ensure the consistency of the labeling
and to account for context~\cite{he2008learning,
  Russell09hierarchical, Gould2009, kumar2010, munoz-10,
  Tighe2010}. Most methods rely on a pre-segmentation into
super-pixels or other segment 
candidates~\cite{Gould2009,kumar2010,munoz-10,Tighe2010}, and extract features
and categories from individual segments and from various combinations
of neighboring segments. The graphical model inference pulls
out the most consistent set of segments that cover the image.

Socher \etal~\cite{SocherEtAl2011} propose a method to aggregate segments in a
greedy fashion using a trained scoring function. The originality of
the approach is that the feature vector of the combination of two
segments is computed from the feature vectors of the individual
segments through a trainable function. Like us, they use ``deep
learning'' methods to train their feature extractor. But unlike us,
their feature extractor operates on hand-engineered features.

One of the main question in scene parsing is how to take a wide
context into account to make a local decision. Munoz \etal~\cite{munoz-10}
proposed to use the histogram of labels extracted from a
coarse scale as input to the labeler that look at finer scales.  Our
approach is somewhat simpler: our feature extractor is applied densely
to an image pyramid. The coarse feature maps thereby generated are
upsampled to match that of the finest scale. Hence with three scales,
each feature vector has multiple fields which encode multiple regions
of increasing sizes and decreasing resolutions, centered on the same
pixel location.

Like us, a number of authors have used trees to generate candidate
segments by aggregating elementary segments, as 
in~\cite{Russell09hierarchical}. Using trees allows to rely on
fast inference algorithms based on graph cuts or other methods. In
this paper, we use an innovative method based on finding a set of
tree nodes that cover the images while minimizing some criterion.

Our system extracts features densely from a multiscale pyramid of
images using a convolutional network (ConvNet)~\cite{lecun-98}.
ConvNets can be fed with raw pixels and can automatically learn
low-level and mid-level features, alleviating the need for
hand-engineered features. One big advantage of ConvNets is the ability
to compute dense features efficiently over large images.  ConvNets are
best known for their applications to detection and
recognition~\cite{osadchy-07,jarrett-iccv-09}, but they have also been
used for image segmentation, particularly for biological image
segmentation~\cite{ning-05,jain-iccv-07,turaga2009}.

The only published work on using ConvNets for scene parsing is that of
Grangier \etal~\cite{Grangier2009}. While somewhat preliminary, their
work showed that convolutional networks fed with raw pixels could be
trained to perform scene parsing with decent accuracy.
Unlike~\cite{Grangier2009} however, our system uses a boundary-based
over-segmentation to align the labels produced by the ConvNet to the
boundaries in the image. Our system also takes advantage of the
boundary-based over-segmentation to produce representations that are
independent of the size of the segment through feature pooling.

\section{An end-to-end trainable model for scene parsing\label{sec:Segmentation-and-recognition} }

The model proposed in this paper, depicted on Figure~\ref{fig:model}, 
relies on two complementary image representations. 
In the first representation, the image is
seen as a point in a high-dimensional space, and we seek to find a
transform $f:\mathbb{R}^P \rightarrow \mathbb{R}^Q$
that maps these images into a space in which each pixel can be assigned
a label using a simple linear classifier. This first representation
typically suffers from two main problems: (1) the window considered
rarely contains an object that is properly centered and scaled, and
therefore offers a poor observation basis to predict the class of
the underlying object, (2) integrating a large context involves increasing
the grid size, and therefore the dimensionality $P$ of the input;
given a finite amount of training data, it is then necessary to enforce
some invariance in the function $f$ itself. This is usually achieved
by using pooling/subsampling layers, which in turn degrades the ability
of the model to precisely locate and delineate objects. In this paper,
$f$ is implemented by a multiscale convolutional network, which allows
integrating large contexts (as large as the complete scene) into local
decisions, yet still remaining manageable in terms of parameters/dimensionality.
This multiscale model, in which weights are shared across scales,
allows the model to capture long-range interactions, without
the penalty of extra parameters to train. This model is described in 
Section~\ref{sub:Scale-invariant,-scene-level-fea}.

In the second representation, the image is seen as an edge-weighted
graph, on which a hierarchy of segmentations/clusterings can be constructed.
This representation yields a natural abstraction of the original pixel
grid, and provides a hierarchy of observation levels for all the objects
in the image. It can be used as a solution to the first problem exposed
above: assuming the capability of assessing the quality of all the
components of this hierarchy, a system can automatically choose its
components so as to produce the best set of predictions. Moreover,
these components are spatially accurate, and naturally delineate the
underlying objects, as this representation conserves pixel-level 
precision. Section~\ref{sec:Paramfree-Parsing} describes our 
methodology.

\subsection{Scale-invariant, scene-level feature extraction\label{sub:Scale-invariant,-scene-level-fea}}

Our feature extractor is based on a convolutional network. Convolutional
networks are natural extensions of neural networks,
in which weights are replicated over space, or in other terms the
linear transforms are done using 2D convolutions. A convolution can
be seen as a linear transform with shared (replicated) weights. The
use of weight sharing is justified by the fact that image statistics
are stationary, and features and combinations of features that are
relevant in one region of an image are also relevant in other regions.
In fact, by enforcing this constraint, each layer of a convolutional
network is explicitly forced to model features that are shift-equivariant. 
Because of the imposed weight-sharing, convolutional networks have
been used successfully for a number of image labeling problems.

More holistic tasks, such as full-scene understanding (pixel-wise
labeling, or any dense feature estimation) require the system to model
complex interactions at the scale of complete images, not simply within
a patch. In this problem the dimensionality becomes unmanageable:
for a typical image of $256\times256$ pixels, a naive neural network
would require millions of parameters, and a naive convolutional network
would require filters that are unreasonably large to view enough context. 

Our multiscale convolutional network overcomes these limitations by
extending the concept of weight replication to the scale space. Given
an input image $\mathbf{I}$, a multiscale pyramid of images 
$\mathbf{X}_{s}, \; \forall s\in\{1,\dots,N\}$
is constructed, with $\mathbf{X}_{1}$ being the size of $\mathbf{{I}}$. The multiscale
pyramid can be a Laplacian pyramid, and is typically pre-processed,
so that local neighborhoods have zero mean and unit standard deviation.
Given a classical convolutional network $f_{s}$ with parameters $\theta_{s}$,
the multiscale network is obtained by instantiating one network per
scale $s$, and sharing all parameters across scales: 
$\theta_{s}=\theta_{0},\:\forall s\in\{1,\dots,N\}$. 

More precisely, the output features are computed using the 
scaling/normalizing function $g_{s}$ as 
$\mathbf{X}_{s}=g_{s}(\mathbf{I})$ for all $s\in\{1,\dots,N\}$.
The convolutional network $f_{s}$ can then be described as a sequence
of linear transforms, interspersed with non-linear symmetric squashing
units (typically the $\tanh$ function):
$\mathbf{F}_{s}=f_{s}(\mathbf{X}_{s};\theta_{s})=\mathbf{W}_{L}\mathbf{H}_{L-1}$,
with
$\mathbf{H}_{l}=\tanh(\mathbf{W}_{l}\mathbf{H}_{l-1}+\mathbf{b}_{l})$ for 
all $l\in\{1,\dots,L-1\}$, where $\mathbf{H}_{l}$ is the vector of hidden units at layer $l$,
for a network with $L$ layers, $\mathbf{H}_{0} = \mathbf{X}_{s}$ 
and $\mathbf{b}_{l}$ is a vector of
bias parameters. The matrices $\mathbf{W}_{l}$ are Toeplitz matrices,
and therefore each hidden unit vector $\mathbf{H}_{l}$ can be expressed
as a regular convolution between the kernel $\mathbf{w}_{lpq}$ and
the previous hidden unit vector $\mathbf{H}_{l-1}$
\begin{equation}
\mathbf{H}_{lp}=\tanh(b_{lp}+\sum_{q\in\mathrm{parents}(p)}\mathbf{w}_{lpq}*\mathbf{H}_{l-1,q}).
\end{equation}

The filters $\mathbf{w}_{lpq}$ and the biases $\mathbf{b}_{l}$ constitute
the trainable parameters of our model, and are collectively denoted
$\theta_{s}$.

Finally, the output of the $N$ networks are upsampled and concatenated so 
as to produce $\mathbf{F}$, a map of feature vectors the size of 
$\mathbf{F}_{1}$, which can be seen
as local patch descriptors and scene-level descriptors
\begin{equation}
\mathbf{F}=[\mathbf{F}_{1}, u(\mathbf{F}_{2}), \dots, u(\mathbf{F}_{N})],
\end{equation}
\noindent
where $u$ is an upsampling function.

As mentioned above, weights are shared between networks $f_{s}$.
Intuitively, imposing complete weight sharing across scales is a natural
way of forcing the network to learn scale invariant features, and
at the same time reduce the chances of over-fitting. The more scales
used to jointly train the models $f_{s}(\theta_{s})$ the better
the representation becomes for all scales. Because image content is,
in principle, scale invariant, using the same function to extract
features at each scale is justified. In fact, we observed a 
performance decrease when removing the weight-sharing.

\subsection{Parameter-free hierarchical parsing}\label{sec:Paramfree-Parsing}

Predicting the class of a given pixel from its own feature vector is
difficult, and not sufficient in practice.
The task is easier if we consider a spatial grouping of feature vectors
around the pixel, \ie a neighborhood. Among all possible neighborhoods,
one is the most suited to predict the pixel's class.
In Section~\ref{sec:Optimal-Cover} we
propose to formulate the search for the most adapted neighborhood as an 
optimization problem. The construction of the cost function that is 
minimized is then described in Section~\ref{sec:Producing-Costs}.

\subsubsection{Optimal purity cover}\label{sec:Optimal-Cover}

\comment{
We will first describe the problem we wish to solve, and then present
a practical approximation of it. Let $R_{k}$ be a partition of image
$\mathbf{I}$ in disjoint neighboring pixels groups. Given an image of size
$w\times h$, there exist ?? unique partitions. Assuming a function
$c$ that can predict the class distribution ${\bf d}_{ki}$ of the
$i$th component of $R_{k}$, as well as a confidence cost $S_{ki}$,
the best segmentation is the one in which the average of these costs
is minimal. More formally, the optimal partition is given by

\begin{equation}
R^{*}=\underset{R_{k}}{\mbox{argmin}}\frac{1}{N_{k}}\underset{i\in\mathrm{comps}(R_{k})}{\sum}S_{ki},
\label{eq:problem}
\end{equation}

\noindent
where $N_{k}$ is the number of components in the $k$th partition.

In practice, the set of partitions $R_{k}$ is too large, and
only a subset of it can be considered. In the next sections, we describe
(1) an efficient method to explore a much smaller subset, (2) a methodology
to produce the confidence costs for all the components considered, and
(3) an efficient procedure to find the best partition in this reduced
subset.
}

We define the neighborhood of a pixel as a connected component that 
contains this pixel. Let $C_{k}, \; \forall k \in \{1,\dots,K\}$ 
be the set of all possible connected components of 
the lattice defined on
image $I$, and let $S_{k}$ be a cost associated to each of these components. 
For each pixel $i$, we wish to find the index $k^{*}(i)$ of the component 
that best explains this pixel, that is, the component with the minimal cost 
$S_{k^{*}(i)}$:
\begin{equation}
k^{*}(i) = \underset{k \;|\; i \in C_{k}}{\mbox{argmin}} S_{k}
\label{eq:problem}
\end{equation}

Note that components $C_{k^{*}(i)}$ are non-disjoint sets that
form a cover of the lattice. Note also that the overall cost 
$S^{*} = \sum_{i} S_{k^{*}(i)}$ is minimal.

\begin{figure}
\begin{centering}
\includegraphics[width=0.85\columnwidth]{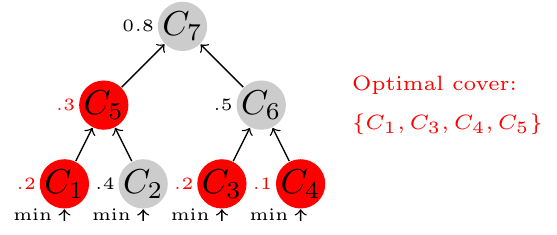}
\par\end{centering}

\caption{Finding the optimal cover. For each pixel (leaf) $i$, the optimal
component $C_{k^{*}(i)}$ is the one along the path between the leaf and the root 
with minimal cost $S_{k^*(i)}$. The optimal cover is the union of
all these components. In this example, the optimal cover 
$\{C_{1},C_{3},C_{4},C_{5}\}$ will result in a segmentation in disjoint sets
$\{C_{1},C_{2},C_{3},C_{4}\}$, with the subtle difference that component 
$C_{2}$ will be labelled with the class of $C_{5}$, as $C_{5}$ is the best
observation level for $C_{2}$.
\label{fig:cut-tree}}
\end{figure}

In practice, the set of components $C_{k}$ is too large, and only a
subset of it can be considered. A classical technique to reduce the
set of components is to consider a hierarchy of
segmentations~\cite{NS96,Arbelaez2010,GuiguesCM06}, that can be
represented as a tree $T$. Solving Eq~\ref{eq:problem} on $T$ can be
done simply by exploring the tree in a depth-first search manner, and
finding the component with minimal weight along each branch.
Figure~\ref{fig:cut-tree} illustrates the procedure.

\comment{
Non-horizontal cuts of the hierarchy can then be obtained using various 
filtering techniques, such as the one presented
in \cite{Morris1986,Felzenszwalb2004}, which aims at producing components 
that are neither too small nor too large. Techniques of that kind typically
rely on arbitrary criterions. 
}

\subsubsection{Producing the confidence costs}\label{sec:Producing-Costs}

\begin{figure}
\begin{centering}
\includegraphics[width=0.99\columnwidth]{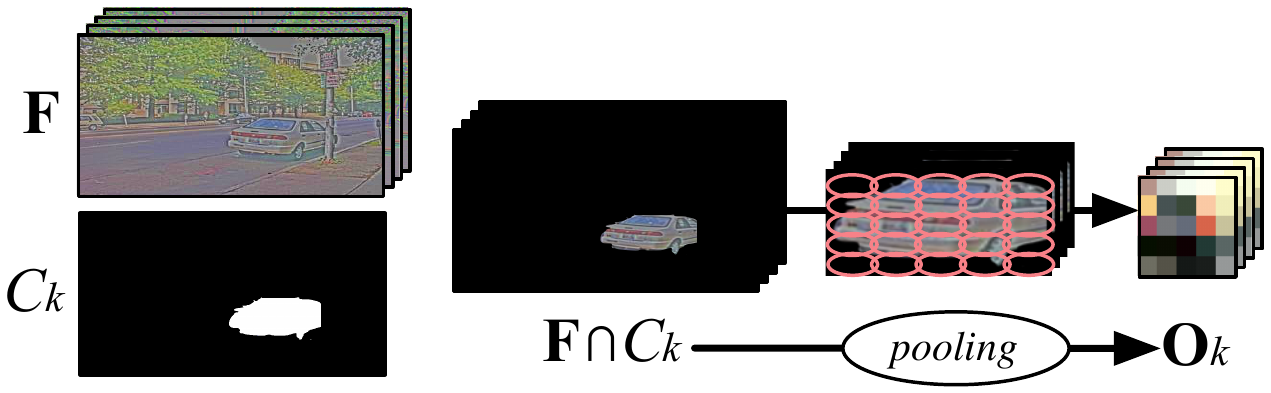}
\par\end{centering}

\caption{The shape-invariant attention function $a$. For each 
component $C_{k}$ in the segmentation tree $T$, the corresponding
image segment is encoded by a spatial grid of feature vectors that 
fall into this segment. The aggregated feature vector of each
grid cell is computed by a component-wise max pooling of the feature
vectors centered on all the pixels that fall into the grid cell;
this produces a scale-invariant representation of the segment and
its surroundings. The result, $\mathbf{O}_{k}$, is a descriptor 
that encodes spatial relations between the underlying object's parts. 
The grid size was set to $5\times5$ for all our experiments.
\label{fig:max-sampler}}
\end{figure}

Given a set of components $C_{k}$, we explain how to produce 
all the confidence costs $S_{k}$. These
costs represent the class purity of the associated components. Given
the groundtruth segmentation, we can compute the cost as being the
entropy of the distribution of classes present in the component. At
test time, when no groundtruth is available, we need to define a
function that can predict this cost by simply looking at the 
component. We now describe a way of achieving this, 
as illustrated in Figure~\ref{fig:max-sampler}.

Given the scale-invariant features $\mathbf{F}$, we define a compact
representation to describe objects as an elastic spatial arrangement
of such features. In other terms, an object, or category in general,
can be best described as a spatial arrangement of features, or parts.
A simple attention function $a$ is used to mask the feature vector
map with each component $C_{k}$, producing a set of $K$
masked feature vector patterns 
$\{\mathbf{F}\bigcap C_{k}\},\;\forall k\in\{1,\dots,K\}$.
The function $a$ is called an attention function because it suppresses
the background around the component being analyzed. The patterns 
$\{\mathbf{F}\bigcap C_{k}\}$
are resampled to produce fixed-size representations. In our model
the sampling is done using an elastic max-pooling function, which
remaps input patterns of arbitrary size into a fixed $G\times G$
grid. This grid can be seen as a highly invariant representation that
encodes spatial relations between an object's attributes/parts. This
representation is denoted $\mathbf{O}_{k}$. Some nice properties
of this encoding are: (1) elongated, or in general ill-shaped objects,
are nicely handled, (2) the dominant features are used to represent
the object, combined with background subtraction, the features pooled
represent solid basis functions to recognize the underlying object.

Once we have the set of object descriptors ${\bf O}_{k}$, we define a
function $c:{\bf O}_{k} \rightarrow [0,1]^{N_{c}}$ 
(where $N_{c}$ is the number of classes) as
predicting the distribution of classes present in component $C_{k}$. 
We associate a cost $S_{k}$ to this distribution. In this paper $c$ is
implemented as a simple 2-layer neural network, and $S_{k}$ is the entropy
of the predicted distribution. More formally, let $\mathbf{x}_{k}$
be the feature vector associated with component ${\bf O}_{k}$, $\hat{{\bf d}}_{k}$
the predicted class distribution, and $S_{k}$ the cost associated
to this distribution. We have
\begin{eqnarray}
\mathbf{y}_{k} & = & {\bf W}_{2}\tanh({\bf W_{\mathbf{\mathrm{1}}}}\mathbf{x}_{k}+{\bf b}_{1}), \label{eq:mlp} \\
\mathbf{\hat{d}}_{k,a} & = & \frac{e^{\mathbf{y}_{k,a}}}{\sum_{b\in\mathrm{classes}} e^{\mathbf{y}_{k,b}} }, \label{eq:dhat} \\
S_{k} & = & -\underset{a\in\mathrm{classes}}{\sum}\mathbf{\hat{d}}_{k,a}\log(\hat{\mathbf{d}}_{k,a}). \label{eq:cost}
\end{eqnarray}
Matrices $\mathbf{W}_{1}$ and $\mathbf{W}_{2}$ are noted $\mathbf{\theta}_{c}$,
and represent the trainable parameters of $c$. These parameters need
to be learned over the complete set of hierarchies, computed on the
entire training set available. The exact training procedure is described
in Section \ref{sec:Training-the-model}.

\section{Training procedure\label{sec:Training-the-model}}

Let ${\cal F}$ be
the set of all feature maps in the training set, and ${\cal T}$ the set of 
all hierarchies.
Training the model described in Section \ref{sec:Segmentation-and-recognition}
can be done in two steps. First, we train the low-level feature extractor
$f_{s}$ in complete independence of the rest of the model. The goal
of that first step is to produce features $({\bf F})_{{\bf F}\in {\cal F}}$ 
that are maximally discriminative for pixelwise classification. 
Next, we construct the hierarchies $(T)_{T \in {\cal T}}$ on the entire training set,
and, for all $T \in {\cal T}$ train the classifier $c$ to predict 
the distribution of classes
in component $C_{k} \in T$, as well as the costs $S_{k}$. 
Once this second part is done, all
the functions in Figure~\ref{fig:model} are defined, and inference
can be performed on arbitrary images. In the next two sections we
describe these two steps.

\subsection{Learning discriminative scale-invariant features}

As described in Section~\ref{sub:Scale-invariant,-scene-level-fea},
feature vectors in ${\bf F}$ are obtained by concatenating the outputs
of multiple networks $f_{s}$, each taking as input a different image
in a multiscale pyramid.
Ideally a linear classifier should produce the correct categorization
for all pixel locations $i$, from the feature vectors $\mathbf{F}_{i}$.
We train the parameters $\mathbf{\theta}_{s}$ to achieve this goal.
Let $\mathbf{c}_{i}$ be the true target vector for pixel
$i$ and $\hat{\mathbf{c}_{i}}$ be the normalized prediction from the 
linear classifier, we set:
\begin{eqnarray}
L_{\mathrm{cat}} & = & \sum_{i\in\mathrm{pixels}}l_{\text{cat}}(\hat{\mathbf{c}_{i}},\mathbf{c}_{i})
\label{eq:l_cat},\\
l_{\mathrm{cat}}(\hat{\mathbf{c}_{i}},\mathbf{c}_{i}) & = & -\sum_{a\in\text{classes}}\mathbf{c}_{i,a}\ln(\hat{\mathbf{c}}_{i,a})
\label{eq:crossent},\\
\hat{\mathbf{c}}_{i,a} & = & \frac{e^{\mathbf{w}_{a}^{T}\mathbf{F}_{i}}}{\sum_{b\in\mathrm{classes}}e^{\mathbf{w}_{b}^{T}\mathbf{F}_{i}}}.
\label{eq:softmax}
\end{eqnarray}
The elementary loss function $l_{\mathrm{cat}}(\hat{\mathbf{c}_{i}},\mathbf{c}_{i})$ 
in Eq \ref{eq:l_cat} is chosen to penalize the deviation of the multiclass
prediction $\hat{\mathbf{c}}_{i}$ from the target vector $\mathbf{c}_{i}$. In this paper,
we use the multiclass \emph{cross entropy} loss function. In order
to use this loss function, we compute a normalized predicted probability
distribution over classes $\hat{\mathbf{c}}_{i,a}$ using the \emph{softmax}
function in Eq \ref{eq:softmax}. The cross entropy between the predicted
class distribution and the target class distribution at a pixel location
$i$ is then measured by Eq \ref{eq:crossent}. The true target probability
$\mathbf{c}_{i,a}$ of class $a$ to be present at location $i$ can either
be a distribution of classes at location $i$, in a given neighborhood
or a hard target vector: $\mathbf{c}_{i,a}=1$ if pixel $i$ is labeled $a$,
and $0$ otherwise. For training
maximally discriminative features, we use hard target vectors in this
first stage. Once the parameters $\mathbf{\theta}_{s}$ are trained, we
discard the classifier in Eq~\ref{eq:softmax}.

\subsection{Teaching a classifier to find its best observation level}

Given the trained parameters $\theta_{s}$, we build ${\cal F}$ and
${\cal T}$, \ie we compute all the vector maps ${\bf F}$ and the hierarchies $T$ 
on all the training data available, so as to produce a new training set of descriptors
$\mathbf{O}_{k}$. This time, the parameters $\theta_{c}$ of the classifier $c$
are trained to minimize the KL-divergence between the true
(known) distributions of labels $\mathbf{d}_{k}$ in each component,
and the prediction from the classifier $\hat{\mathbf{d}}_{k}$
(Eq \ref{eq:dhat}):
\begin{equation}
l_{div}=\sum_{a\in\text{classes}}\mathbf{d}_{k,a}\mbox{ln}(\frac{\mathbf{d}_{k,a}}{\hat{\mathbf{d}}_{k,a}}).
\end{equation}

In this setting, the groundtruth distributions $\mathbf{d}_{k}$ are
not hard target vectors, but normalized histograms of the labels
present in component $C_{k}$. Once the parameters $\theta_{c}$
are trained, $\hat{\mathbf{d}}_{k}$ accurately predicts the 
distribution of labels, and Eq \ref{eq:cost} can be used
to assign a purity cost to the component.

\section{Experiments\label{sec:Experiments}}

We report results on three standard datasets. 
(1) The Stanford Background
dataset, introduced in \cite{Gould2009} for evaluating methods for 
semantic scene understanding. The dataset contains $715$
images chosen from other existing public datasets so that all the
images are outdoor scenes, have approximately $320 \times 240$ pixels, and
contain at least one foreground object.
We use the evaluation procedure introduced in \cite{Gould2009},
$5$-fold cross validation: $572$ images used for training, and $142$
for testing.
(2) The SIFT Flow dataset, as described in Liu \etal \cite{Liu2009}.
This dataset is composed of $2,688$ images that have been thoroughly
labeled by LabelMe users. Liu \etal \cite{Liu2009} have split this
dataset into $2,488$ training images and $200$ test images and used synonym
correction to obtain $33$ semantic labels. We use this same training/test
split.
(3) The Barcelona dataset, as described in Tighe \etal \cite{Tighe2010},
is derived from the LabelMe subset used in~\cite{russell2007object}. It has $14,871$ training and $279$ test images. The test set consists of street scenes from Barcelona, while the training set ranges in scene types but has no street scenes from Barcelona. 
Synonyms were manually consolidated by~\cite{Tighe2010} to produce $170$ unique labels.

For all experiments, we use a $2$-stage convolutional network.
The input $\mathbf{I}$, a $3$-channel image, is transformed into
a $16$-dimension feature map, using a bank of $16$ $7\times7$ filters
followed by $\tanh$ units; this feature map is then pooled using a $2\times2$
max-pooling layer; the second layer transforms the $16$-dimension
feature map into a $64$-dimension feature map, each component being
produced by a combination of $8$ $7\times7$ filters ($512$ filters),
followed by $\tanh$ units; the map is pooled using a $2\times2$
max-pooling layer; finally the $64$-dimension feature map is transformed
into a $256$-dimension feature map, each component being produced
by a combination of $16$ $7\times7$ filters ($2048$ filters). 

The network is applied to a locally normalized Laplacian pyramid 
constructed on the input image.
For these experiments, the pyramid consists of $3$ rescaled versions
of the input ($N=3$), in octaves: $320\times240$, $160\times120$, $80\times60$. 
All inputs are properly padded, and outputs of each
of the $3$ networks upsampled and concatenated, so as to produce
a $256\times3=768$-dimension feature vector map $\mathbf{F}$. 
The network is trained on all $3$ scales in parallel.

Simple grid-search was performed to find the best learning rate and
regularization parameters (weight decay), using a holdout of 10\% 
of the training dataset for validation. More regularization was necessary
to train the classifier $c$. For both datasets, 
jitter was used to artificially expand the size of the training data, 
and ensure that the features do not overfit some irrelevant biases 
present in the data. Jitter includes: horizontal flipping of all images, 
and rotations between $-8$ and $8$ degrees.

In this paper, the hierarchy used to find the optimal cover is a
simple hierarchy constructed on the raw image gradient, based on a
standard volume criterion~\cite{meyer.najman:segmentation,CouNaj11},
completed by a removal of non-informative small components (less than 
100 pixels).
Classically segmentation methods find a partition of the segments
rather than a cover. Partitioning the segments consists in finding an
optimal cut in the tree (so that each terminal node in the pruned tree
corresponds to a segment). We experimented with a number of graph cut
methods to do so, including
graph-cuts~\cite{FordFulkerson-1955,BoykovJolly2001},
Kruskal~\cite{kruskal56} and Power Watersheds~\cite{CouGraNaj11}, but
the results were systematically worse than with our optimal cover
method.

\begin{figure*}
\begin{centering}
\includegraphics[width=0.98\textwidth]{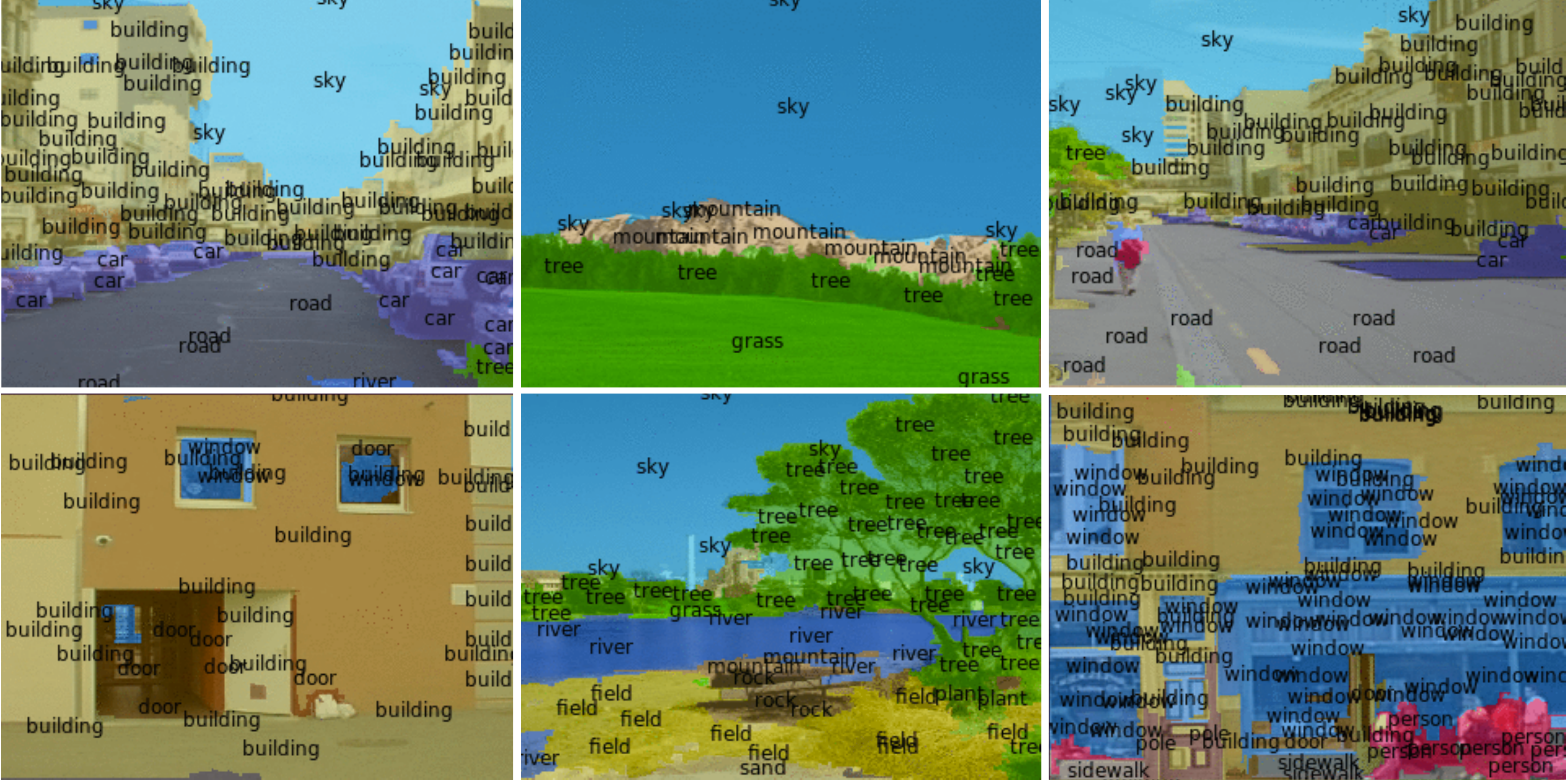}
\par\end{centering}

\caption{Typical results achieved on the SIFT Flow dataset. \label{fig:results}}
\end{figure*}

On the Stanford dataset, we report two experiments: a baseline system, 
based on the multiscale convolutional network alone; and the full model 
as described in Section
\ref{sec:Segmentation-and-recognition}. Results are reported in Table
\ref{tab:results}. On the two other datasets, we report results for our 
complete model only, in Tables~\ref{tab:results-siftflow} 
and~\ref{tab:results-barcelona}. 
Example parses on the SIFT Flow dataset are shown on Figure \ref{fig:results}.

{\bf Baseline, multiscale network:}
for our baseline, the multiscale network is trained as a simple class
predictor for each location $i$, using the single classification
loss $L_{cat}$ defined in Eq \ref{eq:l_cat}. With this simple system,
the pixelwise accuracy is surprisingly good, but the visual aspect
of the predictions clearly suffer from poor spatial consistency,
and poor object delineation.

{\bf Complete system, network and hierarchy:} in this second
experiment, we use the complete model, as described in Section
\ref{sec:Segmentation-and-recognition}. The $2-$layer neural network
(Eq \ref{eq:mlp}) has $3\times3\times3\times256=6912$ input units 
(using a $3\times3$ grid of feature vectors from $\mathbf{F}$), 
$512$ hidden units; and $8$ output units are needed for the Stanford
Background dataset, $33$ for the SIFT Flow dataset, and $170$ for the
Barcelona dataset. Results are significantly better than the baseline
method, in particular, much better delineation is achieved.

For the SIFT Flow dataset, we experimented with two sampling methods
when learning the multiscale features: respecting natural frequencies
of classes, and balancing them so that an equal amount of each class
is shown to the network. Both results are reported in 
Table~\ref{tab:results-siftflow}. Training with balanced frequencies allows
better discrimination of small objects, and although it decreases
the overall pixelwise accuracy, it is more correct from a recognition
point of view. Frequency balancing was used on the Stanford Background
dataset, as it consistently gave better results. For the Barcelona dataset, 
both sampling methods were used as well,
but frequency balancing worked rather poorly in that case. This could
be explained by the fact that this dataset has a large amount of
classes with very few training examples. These classes are therefore
extremely hard to model, and overfitting occurs much faster than
for the SIFT Flow dataset. Results are shown on 
Table~\ref{tab:results-barcelona}.

Results in Table \ref{tab:results} also demonstrate the impressive
computational advantage of convolutional networks over competing algorithms.
Training time is also remarkably fast: results on the Stanford Background
dataset were typically obtained in $24$h on a regular server.

\begin{table}
\begin{centering}
\begin{tabular}{|c|c|c|}
\cline{2-3} 
\multicolumn{1}{c|}{} & P / C & CT\tabularnewline
\hline 
Gould \etal \cite{Gould2009} & 76.4\% / - & 10s to 10min\tabularnewline
\hline 
Munoz \etal \cite{munoz-10} & 76.9\% / 66.2\% & 12s\tabularnewline
\hline 
Tighe \etal \cite{Tighe2010} & 77.5\% / - & 10s to 5min\tabularnewline
\hline 
Socher \etal \cite{SocherEtAl2011} & 78.1\% / - & ?\tabularnewline
\hline 
Kumar \etal \cite{kumar2010} & 79.4\% / - & 10s to 10min\tabularnewline
\hline 
multiscale net & 77.5 \% / 70.0\% & 0.5s\tabularnewline
\hline 
multiscale net + cover & \textbf{79.5\% / 74.3\%} & \textbf{1s}\tabularnewline
\hline 
\end{tabular}
\par\end{centering}

\caption{Performance of our system on the Stanford Background dataset \cite{Gould2009}:
per-pixel accuracy / average per-class accuracy. The third column
reports approximate compute times, as reported by the authors. Note:
we benchmarked our algorithms using a modern $4$-core Intel i7, which
could give us an unfair advantage over the competition. \label{tab:results}}
\end{table}

\begin{table}
\begin{centering}
\begin{tabular}{|c|c|}
\cline{2-2} 
\multicolumn{1}{c|}{} & P / C\tabularnewline
\hline 
Liu \etal \cite{Liu2009} & 74.75 \% / -\tabularnewline
\hline 
Tighe \etal \cite{Tighe2010} & 76.9 \% / 29.4 \%\tabularnewline
\hline 
multiscale net + cover$^{1}$ & \textbf{78.5 \% / 29.6 \%}\tabularnewline
\hline 
multiscale net + cover$^{2}$ & 74.2 \%\textbf{ / 46.0 \%}\tabularnewline
\hline 
\end{tabular}
\par\end{centering}

\caption{Performance of our system on the SIFT Flow dataset \cite{Liu2009}:
per-pixel accuracy / average per-class accuracy. Our multiscale network
is trained using two sampling methods: $^{1}$natural frequencies, 
$^{2}$balanced frequencies. \label{tab:results-siftflow}}
\end{table}

\begin{table}
\begin{centering}
\begin{tabular}{|c|c|}
\cline{2-2} 
\multicolumn{1}{c|}{} & P / C\tabularnewline
\hline 
Tighe \etal \cite{Tighe2010} & 66.9 \% / 7.6 \%\tabularnewline
\hline 
multiscale net + cover$^{1}$ & \textbf{67.8 \% / 9.5 \%}\tabularnewline
\hline 
multiscale net + cover$^{2}$ & 39.1 \% \textbf{10.7} \%\tabularnewline
\hline 
\end{tabular}
\par\end{centering}

\caption{Performance of our system on the Barcelona dataset \cite{Tighe2010}:
per-pixel accuracy / average per-class accuracy. Our multiscale network
is trained using two sampling methods: $^{1}$natural frequencies, 
$^{2}$balanced frequencies. \label{tab:results-barcelona}}
\end{table}

\section{Discussion}

We introduced a discriminative framework for learning to identify
and delineate objects in a scene. Our model does not rely on
engineered features, and uses a multi-scale convolutional network
operating on raw pixels to learn appropriate low-level and mid-level
features. The convolutional network is trained in supervised mode to
directly produce labels. Unlike many other scene parsing systems that
rely on expensive graphical models to ensure consistent labelings, our
system relies on a segmentation tree in which the nodes (corresponding
to image segments) are labeled with the entropy of the distribution
of classes contained in the corresponding segment. Instead of graph
cuts or other inference methods, we use the new concept of {\em
  optimal cover} to extract the most consistent segmentation from the
tree. 

The complexity of each operation is linear in the number of pixels,
except for the production of the tree, which is quasi-linear (meaning
cheap in practice). The system produces state-of-the-art accuracy on the
Stanford Background, SIFT Flow, and Barcelona datasets (both measured
per pixel, or averaged per class), while dramatically outperforming
competing models in inference time.

Our current system relies on a single
segmentation tree constructed from image gradients, and implicitly
assumes that the correct segmentation is contained in the tree. Future
work will involve searches over multiple segmentation trees, or will
use other graphs than simple trees to encode the possible
segmentations (since our optimal cover algorithm can work from other
graphs than trees). Other directions for improvements include the use
of structured learning criteria such as Turaga \etal's Maximin
Learning~\cite{turaga2009} to learn low-level feature vectors from
which better segmentation trees can be produced.

{\small
\bibliographystyle{ieee}
\bibliography{clement,watershed}
}

\end{document}